\documentclass{article}

\usepackage[preprint]{neurips_2025}

\newtheorem{proof}{\bf Proof}
\newtheorem{lemma}{\bf Lemma}
\newtheorem{assumption}{\bf Assumption}
\newtheorem{theorem}{\bf Theorem}
\newtheorem{remark}{\bf Remark}

\usepackage[utf8]{inputenc} 
\usepackage[T1]{fontenc}    
\usepackage{hyperref}       
\usepackage{url}            
\usepackage{booktabs}       
\usepackage{amsfonts}       
\usepackage{amsmath,amsfonts, bm}
\usepackage{newtxmath}
\usepackage{nicefrac}       
\usepackage{microtype}      
\usepackage{xcolor}         

\DeclareUnicodeCharacter{2212}{-}
\DeclareMathOperator{\BR}{\mathbb R}
\DeclareMathOperator{\col}{\mathrm{col}}

\usepackage{graphicx} 
\usepackage{wrapfig}  
\usepackage[skip=6pt]{caption}
\title{Distributed Nash Equilibrium Seeking Algorithm in Aggregative Games for Heterogeneous Multi-Robot Systems}

\author{%
  Yi Dong\\
  University of Liverpool\\
  \texttt{yi.dong@liverpool.ac.uk}\\
  \And
  Zhongguo Li \\
The University of Manchester \\
  \texttt{zhongguo.li@manchester.ac.uk} \\
  \AND
  Sarvapali D. Ramchurn \\
  University of Southampton \\
  \texttt{sdr1@soton.ac.uk} \\
  \And
  Xiaowei Huang \\
  University of Liverpool \\
  \texttt{xiaowei.huang@liverpool.ac.uk} \\
}

\begin{document}

\maketitle

\begin{abstract}
  This paper develops a distributed Nash Equilibrium seeking algorithm for heterogeneous multi-robot systems. The algorithm utilises distributed optimisation and output control to achieve the Nash equilibrium by leveraging information shared among neighbouring robots.
Specifically, we propose a distributed optimisation algorithm that calculates the Nash equilibrium as a tailored reference for each robot and designs output control laws for heterogeneous multi-robot systems to track it in an aggregative game.
We prove that our algorithm is guaranteed to converge and result in efficient outcomes. The effectiveness of our approach is demonstrated through numerical simulations and empirical testing with physical robots.
\end{abstract}

\section{Introduction}
Multi-agent systems involve interactions and decision-making over multiple intelligent agents or entities \cite{ge2018survey}. In such systems, each agent pursues its own interests while being influenced by the actions of other agents. Game theory provides a theoretical framework for analysing this interactive behaviour, and Nash equilibrium is a key concept in game theory, referring to a strategy combination in which no player can improve their outcomes by unilaterally changing their strategies \cite{mas1995microeconomic}. However, Nash Equilibrium Seeking (NES) with a distributed multi-agent system is a complex problem \cite{ye2023distributed,ZHAO2023110933}. 

For instance, in the case of wildfire response and containment operations \cite{queralta2020collaborative,seraj2022multi}, they utilise a heterogeneous team of autonomous agents, including ground robots programmed to extinguish the fire and drones for aerial surveillance. Each robot and drone is programmed not only for cooperative task completion but also for self-preservation. 
Ground robots must strategically position themselves to effectively douse the fire, maintaining a balance between being close enough to combat the flames efficiently and far enough to avoid damage. Meanwhile, drones, capable of gathering critical fire progression data, adjust their flight patterns for optimal data collection and safety. The challenge intensifies as the fire spreads and conditions evolve, requiring real-time strategy adjustments. Each agent must dynamically balance these objectives: the cost of getting too close could be damage to the robot, while staying too far could diminish the effectiveness of containment efforts. Therefore, a strategic equilibrium must be continuously negotiated, ensuring each agent's actions are not only safe but also contribute optimally to the overall objectives of the mission. This dynamic equilibrium adapts as the agents' roles evolve over time, reflecting their changing capabilities and the shifting demands of the firefighting task.
Similar scenarios present significant \textbf{challenges}: \textbf{C1}) The control of robotic agents over a broad area must be addressed, requiring a scalable and robust algorithm for coordination and communication. 
\textbf{C2}) There is an inherent trade-off between the robots' safety and the urgency of rescue operations. An effective strategy must balance maximising rescue efforts with minimising risks to the robots. 
\textbf{C3}) Integrating various types of robotic agents, each with differing system dynamics, into a single operational framework presents significant challenges. These agents have varied behaviours, capabilities, and potential constraints, making it difficult to design a universal control law.

To tackle those challenges, in this paper, we propose a distributed Nash equilibrium seeking algorithm that innovatively employs output regulation techniques to address the dynamic constraints of diverse linear systems\footnote{There is a significant portion of academic research is based on the linearity assumption \cite{huang2016learning,baier2023relinetold}. For broader nonlinear systems, a common simplification is to do the local linearisation \cite{le2024fast}.} while seeking Nash equilibria.
Intuitively, our approach can be divided into two stages. 
In the initial stage (for \textbf{C1} and \textbf{C2}), we leverage the martingale theory \cite{doob1953stochastic} - a mathematical framework used to predict the future behaviours of sequences of random events - to generate a pertinent reference signal and seek a Nash equilibrium. 
In the second stage, we employ output regulation techniques to ensure the asymptotic tracking of a desired reference signal generated from the first stage. 
In contrast to the state-of-the-art approaches \cite{ye2017distributed,salehisadaghiani2016distributed}, which are used to tackle static NES problems, we take the system's intrinsic dynamics into account in the algorithm.
By employing output regulation \cite{ding2022distributed} (for \textbf{C3}) — a control strategy that ensures robotic outputs accurately track desired references even in heterogeneous multi-robot systems — we can effectively address the challenges arising from the system's intrinsic dynamic and guarantee the convergence of the heterogeneous multi-robot systems towards the desired trajectory.
Therefore, the proposed algorithm can guide robots with different dynamics towards their Nash equilibrium points. This algorithm demonstrates the promising potential for applications in heterogeneous multi-robot systems, facilitating the attainment of desirable equilibrium outcomes in various real-world scenarios.

Compared with the existing works, the main contributions of the paper are summarised as follows:

\begin{enumerate}
    \item A distributed Nash equilibrium-seeking algorithm is proposed, and we theoretically prove it converges. Different from most existing studies, e.g., \cite{ren2007information,de2019distributed} and references therein, which are extensively concentrated on continuous-time integrators.
    \item Our approach combines martingale theory and output regulation techniques, specifically designed for heterogeneous linear systems. By integrating these two components, the algorithm enables effective coordination and convergence towards Nash equilibrium points in the presence of heterogeneous dynamics and objectives.
    \item The proposed algorithm has been successfully validated on a real-world multi-robot system, demonstrating its practical applicability. Furthermore, to promote transparency and facilitate reproducibility, all source codes related to the algorithm are openly available on GitHub,
    enabling researchers to validate and further build upon the findings.
\end{enumerate}


\section{Related Work}\label{sec_pre}
In this section, we primarily discuss how the existing state-of-the-art methods address the three challenges mentioned in the introduction. Regarding the first challenge, the existing methods mainly adopt some distributed solutions to reduce system complexity, ensuring the scalability of the system. For the second challenge, most of them formulate the problem as a game and then use some game theory solutions to find the Nash equilibrium points to balance different optimization objectives. For the third challenge, namely the heterogeneous system, most of the existing methods employ end-to-end approaches. By treating system dynamics as a black box, these methods directly process the input from the heterogeneous components and generate the output without delving into the system's dynamics. Or, they simplify the system by formulating it into a single-integrator system.

To address \textbf{C1} and \textbf{C2}, several approaches \cite{tatarenko2020geometric,belgioioso2022distributed,salehisadaghiani2016distributed,ye2017distributed,ye2022distributed} (in general, first-order integrator-based dynamics) have been proposed to solve distributed NES problems, such as, consensus-based methods \cite{ye2017distributed} and gossip-based methods \cite{salehisadaghiani2016distributed}.
Tatarenko \textit{et al.} presented an augmented game mapping to guarantee strong monotonicity, and they also formulated a distributed gradient play algorithm, known as GRANE, aimed at identifying a Nash equilibrium through local information exchange \cite{tatarenko2020geometric}.
Salehisadaghiani and Pavel presented an asynchronous gossip-based algorithm for achieving a Nash equilibrium in a distributed multi-player network, where players base their decisions on estimates of other players' actions from local neighbours \cite{salehisadaghiani2016distributed}. Ye and Hu investigated a distributed NES strategy for noncooperative games in a network with limited player observation, using a leader-following consensus protocol and gradient play to achieve convergence in nonquadratic games with multiple Nash equilibria and global convergence in quadratic games \cite{ye2017distributed}. Gadjov and Pavel proposed a robust distributed algorithm that ensures convergence to the Nash equilibrium by filtering information from observation and communication graphs, addressing challenges posed by noncooperative games with adversarial agents and communication vulnerabilities \cite{gadjov2023algorithm}. 

The aforementioned literature primarily addresses distributed Nash equilibrium but overlooks intrinsic system dynamics (\textbf{C3}). In other words, the objective functions are defined without incorporating dynamic constraints (referring to our subsequent formulation in \eqref{eqn: objective}). This critical difference gives rise to the fundamental principles in designing the NE-seeking algorithms: in the first case, algorithms can take any form, while in the second case, the algorithm has to take the dynamics of the system into account, which makes the studied problem in this paper particularly challenging. To seek Nash equilibria that concurrently satisfy system dynamic constraints, some preliminary research has been undertaken, e.g. \cite{hu2021decentralized,hu2022robust}, which have primarily considered simple single-integrator dynamics. However, in real-world applications, such simple single-integrator systems cannot adequately represent complex system dynamics, and may lead to safety issues due to underlying dynamic behaviours.

\section{Preliminaries}\label{sec_pre}
In this section, we recall previous works that were primarily applied to achieve this paper's main results. $Graph\ Theory$ is applied to describe the communication of the multiagent system. $Nash\ Equilibrium$ is the ultimate goal for this work; the proposal algorithm can make sure the multiagent system will converge to the Nash equilibrium points. $Martingale\ Theory$ is mentioned here to provide a theoretical guarantee to support the following convergence analysis. Finally, section \ref{sec:pf} formally describes the problem researched in this paper.

\textit{Notations:} Let $\mathbb{R}^n$ be the set of vectors with dimension $n>0$. Let $\|x\|$ and $x^T$ be the standard Euclidean norm and the transpose of $x\in\mathbb{R}^n$, respectively. $I_p$ is the identity matrix with dimension $p>0$ and $\otimes$ denotes the Kronecker product.
$\col(\cdot)$ represents the column vector.
\subsection{Graph Theory}\vspace{-6pt}
Following \cite{ren2005consensus}, a directed graph $\mathcal{G(V,E,A)}$ consists of  $\mathcal{V}=\{\nu_{1},\linebreak\cdots,\nu_{N}\}$ as a node set and $\mathcal{E}\in \mathcal{V}\times\mathcal{V}$ as an edge set. If the node $\nu_{i}$ is a neighbour of node $\nu_{j}$, then $(\nu_{i},\nu_{j})\in \mathcal{E}$. A directed graph is strongly connected if there exists a directed path that connects any pair of vertices. $\mathcal{A}$ is the communication matrix and we let $[\mathcal{A}]_{ij}=a_{ij}$ 
where $a_{ij}>0$ if $(\nu_{j},\nu_{i})\in \mathcal{E}$ and $a_{ij}=0$ otherwise. 

\begin{assumption}\label{asm: graph connectivity}
The communication matrix is doubly stochastic, i.e., $\sum_{i}a_{ij}=1, for\ all\ j\in \mathcal V$ and $\sum_{j}a_{ij}=1,  for\ all\ i \in \mathcal V$.
\end{assumption}


\begin{remark}
    If a graph is connected, we can design a communication matrix that satisfies Assumption  \ref{asm: graph connectivity} \cite{nedic2009distributed}. Moreover, each agent allocates a positive weight to the information obtained from its neighbours and assigns a zero weight when no connecting edge exists between agents. Note that Assumption~\ref{asm: graph connectivity} represents a standard connectivity requirement in multi-agent and learning problems \cite{6705625,9029217,7862143}. 
\end{remark}

\subsection{Nash Equilibrium for Aggregative Games}\vspace{-6pt}
In a game involving multiple players, each player aims to maximise their own payoff or utility \cite{jensen2010aggregative}. Nash equilibrium occurs when, given the strategies chosen by other players, no player can unilaterally improve their payoff by changing their strategy. It is a stable state where all players' strategies are mutually consistent, and no player has an incentive to deviate from their chosen strategy.

In an aggregative game, e.g. traffic routing \cite{benenati2024probabilistic}, power system scheduling \cite{liu2017optimal} and federated optimisation \cite{wang2022cooperative}, each player's payoff is influenced by both its individual action and a cumulative function of the actions enacted by all players.
Let $h(\boldsymbol{y}) = \frac{1}{N}\sum_{i=1}^N h_i(y_i)$ denote the aggregative function, where $y_i$ denotes the action of the $i$th agent, $h_i(\cdot)$ is the local mapping function from local decision $y_i$ to aggregated term and $\boldsymbol y = \col(y_1,\dots, y_N)$ as the action vector. Additionally, $y_{-i}$ is defined as the collection of actions from all agents except for agent $i$.
Formally, a Nash equilibrium for an aggregative game can be expressed as:
\begin{equation}  \label{NE} \small
    f_i(y_i^*, h_i(y_i^*, y_{-i}^*)) \leq f_i(y_i', h_i(y_i', y_{-i}^*)), \quad \forall i\in \mathcal{V}, \quad \forall y'_i\in\Omega_i,
\end{equation}
where $\Omega_i$ is the local convex and compact decision set, and $\mathcal{V}$ is the set of agents. If \eqref{NE} is satisfied, then a strategy $\boldsymbol{y}^*$ is said to be a Nash equilibrium. Condition \eqref{NE} means that all players simultaneously take their own best (feasible) responses at $y_i^*$, where no player can further decrease its cost function $f_i(\cdot)$ by changing its decision variable unilaterally.



\subsection{Martingale Theory}\vspace{-6pt}
In the context of convergence analysis, super-Martingale theory plays a crucial role in studying the convergence of stochastic processes and sequences of random variables. It provides a framework to analyse the limiting behaviour of sequences of processes. 
In convergence analysis, one is often interested in understanding convergence properties of sequences of random variables or stochastic processes as the sample size or time horizon increases. Super-Martingale provides a natural class of processes that allow for a broad range of behaviours, including both continuous paths and jumps, making them suitable for studying convergence in various settings.

Considering a random sequence ${\Upsilon} = \{\upsilon_0, \cdots, \upsilon_k, \cdots\}$, if $\mathbb E\upsilon_0<\infty$ and $ \mathbb E(\upsilon_{k+1}|\upsilon_0,\cdots,\upsilon_k)\leq \upsilon_k$, then the sequence ${\Upsilon}$ is called a super-Martingale, where  $ \mathbb E(\upsilon_{k+1}|\upsilon_0,\cdots,\upsilon_k)$ is the conditional mathematical expectation of $\upsilon_{k+1}$ for a given $\upsilon_0, \cdots, \upsilon_k$. With the above definition and Robbins-Siegmund theorem \cite{robbins1971convergence,ram2008distributed}, we have the following lemma.
\begin{lemma}\label{lemma:robbins}
    Let $V_k$, $u_k$, $\beta_k$ and $\alpha_k$ be non-negative random variables adapted to some $\sigma$-algebra $\mathcal{F}_k:=\{V_0, \cdots, V_k, u_0, \cdots, u_k, \beta_0, \linebreak\cdots, \beta_k, \text{ and } \alpha_0, \cdots, \alpha_k\}$. If almost surely $\sum_{k=0}^\infty u_k<\infty$, $\sum_{k=0}^\infty \beta_k<\infty$, and
    \begin{align}  
        \mathbb{E}[V_{k+1}|\mathcal{F}_k]\leq(1+u_k)V_k - \alpha_k+\beta_k \quad \forall k\geq 0,
    \end{align}
    then almost surely $V_k$ converges and $\sum_{k=0}^\infty \alpha_k<\infty$.
\end{lemma}

Here, $\mathbb{E}[V_{k+1}|\mathcal{F}_k]$ denotes the conditional mathematical expectation for the given $V_0, \cdots, V_k$, $u_0, \cdots, u_k$, $\beta_0, \cdots, \beta_k$ and $\alpha_0, \cdots, \alpha_k$.

\subsection{Problem Formulation}\label{sec:pf}\vspace{-6pt}
\begin{wrapfigure}{r}{0.5\textwidth}
    \centering
    \vspace{-20pt}
\begin{subequations} \label{eqn: objective}
\begin{align} 
\min_{y_i \in \Omega_i} & f_i(y_i, \bar y) , \forall i\in \mathcal{V}\label{eqn:line-1} \\
\text{s.t. } x_i(k+1) & = A_ix_i(k) +B_i u_i(k)\label{eqn:dynamic-1} \\
y_i(k) & = C_ix_i(k)\label{eqn:dynamic-2}
\end{align}
\end{subequations}
\vspace{-30pt}
\end{wrapfigure}
In an aggregative game, each agent is a selfish unit that aims to optimise its own objective function, that is:

where $y_i \in \Omega_i \subset \BR^q$ denotes the output of the $i$th agent with $\Omega_i$ being the decision space of agent $i$, and $\bar y = h(y) = \frac{1}{N}\sum_{i=1}^{N} y_i$ denotes the aggregate of all agents in the network. 
Here, we consider a group of heterogeneous linear systems competing over a network, of which their dynamics are governed by Equations \eqref{eqn:dynamic-1} and \eqref{eqn:dynamic-2}. $x_i$ denotes the state space of the robots, for instance, in the case of a ground vehicle, $x_i$ could represent the current position signal of the $i$-th robot. 
$A_i$ matrix is the state transition matrix, which describes how the system’s internal states evolve over time in the absence of external input. $B_i$ matrix is the control input matrix, which characterizes how the external input $u_i$ affects the system state $x_i$. $C_i$ matrix is the observation matrix, describing how the internal system state $x_i$ is mapped to the measurable output $y_i$. 
Notice that even the same type of robots can exhibit distinct dynamics, variations may arise from differences in wheel sizes. Consequently, the proposed heterogeneous approach offers more adaptability in practical applications. 

From \eqref{eqn: objective}, it is clear that the objective function $f_i$ of the $i$-th agent not only depends on its own decision variable $y_i$ but also the aggregate of all others' decisions. 
Seeking for the Nash equilibrium becomes the target of the agents. It is clear that there is no agent that can benefit from unilaterally changing its strategy when a Nash equilibrium has been reached.



To maintain the uniqueness and feasibility of the aggregative game, several conditions should be satisfied. To simplify the derivation, we use $F_i(y_i,\bar{y}_i) = \nabla_{y_i} f_{i}(y_i,\bar{y}_{i})$
\begin{assumption}\label{a:smoothness}
    The objective functions $f_i(\boldsymbol{y}), \forall i \in \mathcal V$ are twice continuously differentiable. All the local decision sets, $\Omega_i$ for all $i\in\mathcal V$, are non-empty, compact and convex.
\end{assumption}

\begin{assumption}\label{a:monotonicity}
    The pseudo gradient mapping $\Phi(\bm y)$ is strictly monotone over $\Omega$ with $\Omega: = \Omega_1\times \Omega_2,\dots, \times \Omega_N $, i.e., 
    \begin{align} 
        \left(\Phi(\bm y')-\Phi(\bm y)\right)^T\left(\bm y'-\bm y\right)>0, \forall\ \bm y,\bm y'\in\Omega,\ and\ \bm y\neq \bm y'
    \end{align}
where $\Phi(\bm y) \triangleq \left(F_{1}(y_1,{\bar{y}}_{1}),
\cdots,F_{N}(y_N,\bar{y}_{N})\right)$. 
\end{assumption}

\begin{remark}
    Consider the aggregative Nash game with Assumptions \ref{a:smoothness} and \ref{a:monotonicity} hold, then we can claim that the game admits a unique Nash equilibrium \cite{koshal2016distributed}. For rigorous theoretical guarantees, the above two assumptions are necessary and also widely used in existing studies \cite{ye2017distributed,li2020distributed}. Under some circumstances, Nash equilibrium cannot be achieved through optimisation, e.g. Assumptions \ref{a:smoothness} and \ref{a:monotonicity} may not hold in some robotics and control applications. However, based on Assumptions \ref{a:smoothness} and \ref{a:monotonicity}, within our context, reaching a Nash equilibrium can be achieved through optimisation.
\end{remark}

\begin{assumption}\label{a:lipstchitz}
    Each mapping $F_i(y_i,z)$ is uniformly Lipschitz continuous in $z$ over $\Omega$, for every fixed $y_i \in \Omega_i$ i.e. for some $L_i \geq 0$ and for all $z_1,z_2\in\Omega$
    \begin{align}
        \| F_{i}(y_i,z_1) -  F_{i}(y_i,z_2)\| \leq L_i\|z_1-z_2\|
    \end{align}
\end{assumption}

In many real-world applications, the objective functions in \eqref{eqn: objective} are defined according to the distance between the agent and the position of interest, for example, swarm robots \cite{soria2022distributed}, source seeking \cite{li2021concurrent}, search and rescue \cite{azzollini2021UAV}. 
To elucidate this formulation more clearly, we consider a running example involving multi-agent systems, specifically unmanned ground vehicles. In this scenario, each vehicle's objective is to find a balance between its individual local target and the global midpoint of all vehicles. For instance, in the robotic source-seeking problem, the global midpoint is the radiation centre. Robots need to approach the source to absorb radiation, yet they must not get too close to avoid collisions. Thus, seeking a balance is essential, and it is important to note that any movement by a single vehicle will shift the global midpoint. Furthermore, each vehicle possesses its own unique dynamics, as illustrated in equations \eqref{eqn:dynamic-1} and \eqref{eqn:dynamic-2}. Therefore, the problem can be formulated as 
\begin{align}  \label{eqn:NE}
\begin{aligned}
    f_i(y_i,\bar{y}) = \|y_i - r_i\|^2 + \| y_i - \frac{1}{N}\sum_{j=1}^N {y}_j \|^2 \qquad \ \ 
   \text{s.t. } \ y_i \in \Omega_i.
    \end{aligned}
\end{align}
Equation \eqref{eqn:NE} is an example of \eqref{eqn:line-1}. 
Robots aim to reach a compromise between their individual objective of moving to position $r_i$ and the collective objective of maintaining connectivity with their neighbours. This compromise is reflected in the NE state, where agents choose strategies that balance these two objectives.
In this case, vehicle $i$ intends to minimise the cost function $f_i$, which includes the information of balancing its local target $r_i$ and global midpoint $\frac{1}{N}\sum_{i=1}^N {y}_i$. $\Omega_i$ is designed as the compact set of maps, denoting that vehicles can only run within a certain map.




%
\section{NE Algorithm}\label{sec_algorithm}

\subsection{Algorithm Design}\vspace{-6pt}
To address the NES problem for heterogeneous linear systems, we break down the problem into two sub-problems: Nash equilibrium seeking and tracking of heterogeneous systems. Intuitively, robots in multi-agent systems are usually designed with efficient tracking algorithms to follow instructions given by high-level decision-makers. Here, \eqref{solution:v1} and \eqref{solution:v2} are used to generate such tracking references, where $v_i$ is the message to be communicated. We demonstrate that the proposed algorithm converges to the Nash equilibrium points, as detailed in \ref{parta}. Then, we employ output regulation technology, \eqref{solution:ref1} and \eqref{solution:ref2}, to ensure that heterogeneous systems can accurately track these references. The efficacy of the approach in achieving convergence is shown in \ref{partc}. Furthermore, in constrained optimisation problems, ensuring that the solution remains within the set of constraints is crucial. Therefore, we introduce the projection operator $P_\Omega$ in \eqref{solution:ref2}, which ensures each algorithm iteration meets these constraints, securing a valid final solution.

\begin{wrapfigure}{r}{0.5\textwidth}
    \centering
    \vspace{-20pt}
\begin{subequations}\label{solution:B}\small
\begin{align}  
u_i(k) &= -K_ix_i(k) + (G_i + K_i\Psi_i)\xi_i(k)\label{solution:ref1}\\
\xi_i(k+1) &= P_{\Omega_i}\left(\xi_i(k) - \alpha F_i(\xi_i,\hat{v}_i)\right)\label{solution:ref2}\\
\hat{v}_i(k) &= \sum_{j=1}^N a_{ij}v_j(k)\label{solution:v1}\\
        v_i(k+1) &= \hat{v}_i(k) + \xi_i(k+1) - \xi_i(k)\label{solution:v2}
\end{align}
\end{subequations}
\vspace{-30pt}
\end{wrapfigure}
Therefore, the distributed Nash equilibrium-seeking algorithm for heterogeneous linear systems is designed as \eqref{solution:B}: 

where $\xi_i(k)$ is an internal auxiliary variable generating the tracking reference for the $i$th agent. $P_{\Omega_i}$ represents the project function. $v_i$ is the local estimation of the auxiliary variable of all other agents.
$ K_i $ is chosen such that $A_i-B_iK_i$ is Schur stable under assumption that the dynamics $(A_i,B_i)$ are controllable. 
{$ G_i $ and $\Psi_i$ are gain matrices, which can be obtained by solving the following 
}
\begin{equation}  
\label{eqn: linear matrix align}
		 (A_i-I)\Psi_i + B_iG_i= 0, \quad
		 C_i\Psi_i - I = 0 .
\end{equation}
{Intuitively, the gain matrices $K_i, G_i, \Psi_i$ are designed to track the reference according to system dynamics $A_i, B_i, C_i$. For heterogeneous systems, every robot $i$ may have different $A_i, B_i, C_i$.}
To ensure the solvability of \eqref{eqn: linear matrix align}, we adopt the following Assumption \ref{asm: rank}, 
which is a regulation equation in the output regulation literature \cite{huang2004nonlinear}. 
\begin{assumption}\label{asm: rank}
	The pairs $(A_i,B_i),\forall i \in \mathcal V$ are controllable,
and 
\begin{align}  \label{eqn: rank condition}
	\operatorname{rank}\left[\begin{array}{cc}
		A_i-I & B_i\\
			     C_i & 0 
	\end{array}\right]=n +q .
\end{align}
\end{assumption}

\begin{remark}
In Assumption \ref{asm: rank}, controllability is a typical requirement for control algorithms, and Equation \eqref{eqn: rank condition} is essential to ensure the solvability of Equation \eqref{eqn: linear matrix align}. The proposed algorithm is, in fact, a combination of gradient-descent optimisation and output regulation techniques. The internal model $\xi_i(k)$ is generated by consensus-based gradient-descent optimisation, and $\hat v_i$ is employed to estimate the aggregate term of the game. {The design of the control input $u_i$ is motivated by the classic output regulation approach \cite{huang2004nonlinear}.}
\end{remark}

\subsection{Convergence Analysis} \label{sec:CA}\vspace{-6pt}

The proposed algorithm's convergence analysis is carried out in two stages. Firstly, we establish that algorithm \eqref{solution:B} will converge to the Nash equilibrium of problem \eqref{eqn: objective}. Following this, we demonstrate that the heterogeneous linear systems will also attain the unique optimal Nash equilibrium points.

\textbf{Nash Equilibrium Seeking: }\label{parta}
We begin with the convergence analysis for seeking the NE in algorithm~\eqref{solution:B}, which will then serve as a reference generator later in the proof of linear system regulation. 

\begin{theorem}\label{thm: 1}
Let Assumption~\ref{asm: graph connectivity}-\ref{a:monotonicity} hold, $L_i$ being the Lipschitz constant (cf. \ref{a:lipstchitz}) of the cost function, then algorithm~\eqref{solution:B} converges to the Nash equilibrium  $Y^*$. 
\end{theorem}
To prove the system can finally achieve the optimal points, we introduce the discrete-time Lyapunov function: Let $V(k)= \sum_{i=1}^N\|\xi_i(k) - y_i^*\|^2$, then we have:
\begin{equation}\label{eqn:V}
\small
    \begin{aligned}  
        &V(k+1)= \sum_{i=1}^N\|\xi_i(k+1) - y_i^*\|^2\notag
        =\sum_{i=1}^N\|P_{\Omega_i}\left(\xi_i(k) - \alpha F_{i}(\xi_i(k),\hat{v}_i(k))\right) - y_i^*\|^2\\\notag
        =&\sum_{i=1}^N\|P_{\Omega_i}\left(\xi_i(k) - \alpha F_{i}(\xi_i,\hat{v}_i)\right) - P_{\Omega_i}\left(y_i^* - \alpha F_{i}(y_i^*,\bar{y}^*)\right)\|^2\notag
        \leq\sum_{i=1}^N\|\xi_i(k) - y_i^* - \alpha(F_{i}(\xi_i,\hat{v}_i)  -F_{i}(y_i^*,\bar{y}^*))\|^2\\\notag
        \leq&\sum_{i=1}^N\|\xi_i(k) - y_i^*\|^2 + \alpha^2\sum_{i=1}^N\|\underbrace{ F_{i}(\xi_i,\hat{v}_i)  -F_{i}(y_i^*,\bar{y}^*)}_{\mathcal{M}_i(k)}\|^2 - 2\alpha\sum_{i=1}^N\underbrace{( F_{i}(\xi_i,\hat{v}_i)  -F_{i}(y_i^*,\bar{y}^*))^T(\xi_i(k) - y_i^*)}_{\mathcal{V}_i(k)}\\
        =& V(k) + \alpha^2\sum_{i=1}^N \mathcal{M}_i(k) - 2\alpha \sum_{i=1}^N \mathcal{V}_i(k)\notag
    \end{aligned}
\end{equation}

With this definition, we can finally get that $V(k)$ is asymptotically stable, which implies that $\xi_i(k) \rightarrow y_i^*$. The full process of the proof can be found in Appendix \ref{App:A}.

\textbf{Nash Equilibrium Tracking: }\label{partc}
Based on the convergence result of the Nash equilibrium in Theorem~\ref{thm: 1}, the remaining task is to regulate the heterogeneous dynamic systems to track the Nash equilibrium point. It shall be highlighted that the state-of-the-art works in NES are mostly concentrated on the first mission, while in this paper, we further establish Nash equilibrium seeking and tracking for heterogeneous systems with dynamics in \eqref{eqn:dynamic-1} and \eqref{eqn:dynamic-2}. To achieve this target, we resort to output regulation techniques. 

Now, we apply a state transformation to \eqref{solution:B}, leading to 
\begin{equation}
    x_{i,s}(k)  = \Psi_i \xi_i(k) ,\quad
    u_{i,s}(k)  = G_i \xi_i(k),
\end{equation}
based on which we define the error dynamics as 
\begin{equation}
    \bar x_i(k)  = x_i(k)-x_{i,s}(k), \quad
    \bar u_i(k)  = u_i(k)-u_{i,s}(k),
\end{equation}
Invoking the control design in \eqref{solution:B} leads to the closed-loop error dynamics 
\begin{equation}  \label{eqn: closed-loop error dynamics}
	\bar x_i(k+1)   =  (A_i-B_iK_i)\bar x_i(k) - \Psi_i \Delta_i(k) , \quad
			e_i(k)  = C_i\bar x_i(k). 
\end{equation}
where $\Delta_i(k) = P_{\Omega_i}\left(\xi_i(k) - \alpha F_{\xi_i}(\xi_i(k),\hat{v}_i(k))\right) - \xi_i(k)  $.


Now we are ready to present the overall convergence result of the proposed algorithm in \eqref{solution:B}, where the full process of the proof can be found in Appendix \ref{App:B}:
\begin{theorem}\label{thm: 2}
Let Assumptions~\ref{asm: graph connectivity}-\ref{asm: rank} hold. If $K_i$ is chosen such that $A_i-B_iK_i$ is Schur stable and the $G_i$ and $\Psi_i$ are designed by solving the regulation equations in \eqref{eqn: linear matrix align}, then the proposed algorithm in \eqref{solution:B} solves the Nash equilibrium problem in \eqref{eqn: objective}.
\end{theorem}

With Theorems \ref{thm: 1} and \ref{thm: 2}, we can claim that the proposed algorithm can control the heterogeneous linear systems to the unique optimal Nash equilibrium points.

\section{Experiment Results}\label{sec_sim}
In this experiment, we examine the proposed algorithm through two cases. The first case involves numerical experiments to verify its correctness and demonstrate its generalizability, scalability and robustness across different robot types in simulations. After that, the algorithm is deployed on real-world robots to assess its practicality. 
We conducted all the experiments on a laptop with an Intel i7 chip, a RTX 2080super GPU and 32GB memory.

\subsection{Numerical Simulation}\vspace{-6pt}

\begin{wrapfigure}{r}{0.5\textwidth}
    \centering
    \vspace{-10pt}  
    \includegraphics[width=\linewidth]{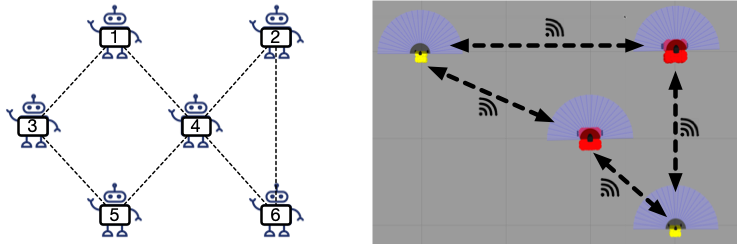}
    \caption{Communication Graphs.}
    \label{fig:Ring}
    \vspace{-10pt}  
\end{wrapfigure}

\begin{figure}[htbp]
    \centering
    \includegraphics[width=\hsize]{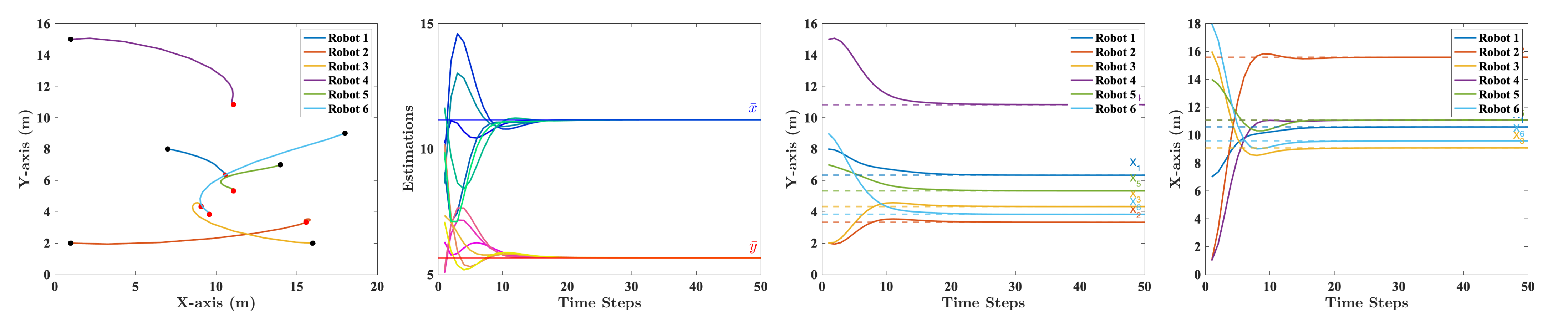}
    \caption{Numerical Simulation Results of Effectiveness.}
    \vspace*{-6pt}
    \label{fig:NSim}
\end{figure}

\textbf{Effectiveness: }In this case, we construct the experiment with six robots to demonstrate the effectiveness of the proposed algorithm. In this scenario, each robot is governed by its own linear dynamics: $A_{1,2} = [0, 1; 0, 0]$, $B_{1,2} = [0, 1; 1, −2]$, $C_{1,2} = [1, 1]$, $A_{3,4} = [0, −1; 1, −2]$, $B_{3,4} = [1, 0; 3, −1]$, $C_{3,4} = [−1, 1]$, $A_{5,6} = [0, 1, 0; 0, 0, 1; 0.5, 1, −2]$, $B_{5,6} = [1, 0; 0, 1; 1, 0]$, $C_{5,6} = [1, −1, 1]$; The initial points are $r_x = [6,1,16,1,14,18], r_y = [8,2,2,15,7,9]$, while their targets are $r_x = [10,18,7,10,11,8], r_y = [6,1,3,17,5,2]$. 
For the purpose of simplification, we assume that the robots possess rigid bodies during this experiment. Specifically, the communication graph is designed as depicted in Fig.  \ref{fig:Ring} (left) and the experimental results are depicted in Fig. \ref{fig:NSim}.

Fig. \ref{fig:NSim} illustrates the Nash equilibrium seeking process of the proposed algorithm for six robots. In the first graph, the black markers represent the starting points, while the coloured lines depict the movement trajectories of the robots. The red markers indicate the endpoints. In the second graph, different coloured lines represent the estimations of the mean value $\hat{v}$ (cold colour lines for ${x_i}$ and hot colour lines for ${y_i}$) by the six robots. To provide clearer visualization, the red line represents the ground truth $\hat{x}$ mean value, and the blue line represents the ground truth $\hat{y}$ mean value. The third and fourth graphs record the trajectories of $x_i$ and $y_i$ for the six robots, where the dashed lines represent the theoretical Nash equilibrium values. It can be observed that the values of $x_i$ and $y_i$ approach steady state and  $\hat{v}$ accurately estimate the aggregate term. 
The results of this experiment demonstrate that our algorithm effectively enables the six robots to estimate the target mean values and ultimately reach a Nash equilibrium state.


\begin{wrapfigure}{r}{0.5\textwidth}
    \centering
    \vspace{-10pt}  
    \includegraphics[width=\hsize]{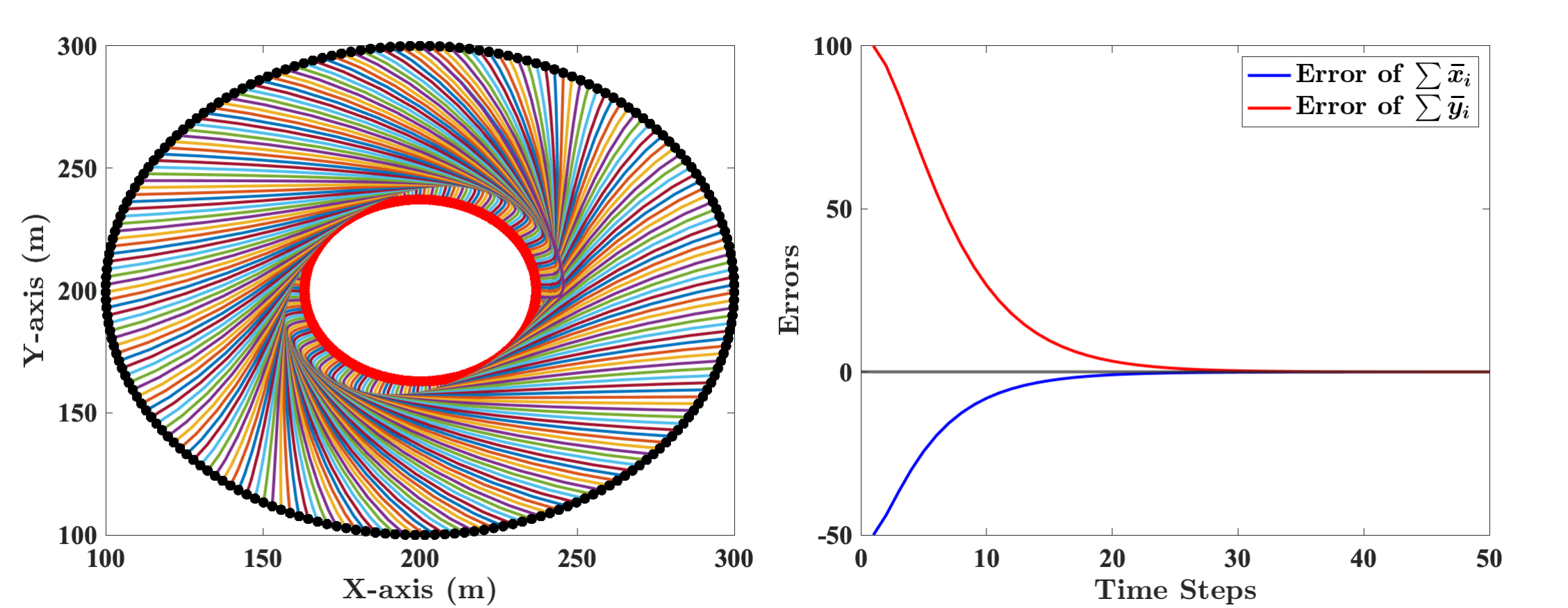}
    \caption{Simulation Results of Scalability.}
    \label{fig:case_L}
    \vspace{-10pt}  
\end{wrapfigure}

\textbf{Scalability: }
In this case, we conduct the experiment to evaluate the scalability of the proposed algorithm. We formed a circular topology consisting of 200 robots, where each robot was only allowed to communicate with its six neighbouring robots. Results are summarised in Fig. \ref{fig:case_L}.


In the left figure of Fig. \ref{fig:case_L}, we can observe that all the robots are heading towards their respective target positions. To provide a clearer demonstration, we summed up the errors for 400 terms ($x_i-x_i^*$, $y_i-y_i^*$, where $x_i^*$ and $y_i^*$ are the corresponding theoretical Nash equilibrium), and the results are shown in the right figure of Fig. \ref{fig:case_L}. From this experiment, we can see that the proposed algorithm is capable of guiding different robots to their corresponding Nash equilibrium points (with errors approaching zero) even in a large-scale network of 200 agents. In the numerical experiments, 6 robots take approximately 0.0029 seconds to reach the Nash equilibrium, while 200 robots take about 0.5242 seconds.

\begin{wrapfigure}{r}{0.5\textwidth}
    \centering
    \vspace{-10pt}  
    \includegraphics[width=\hsize]{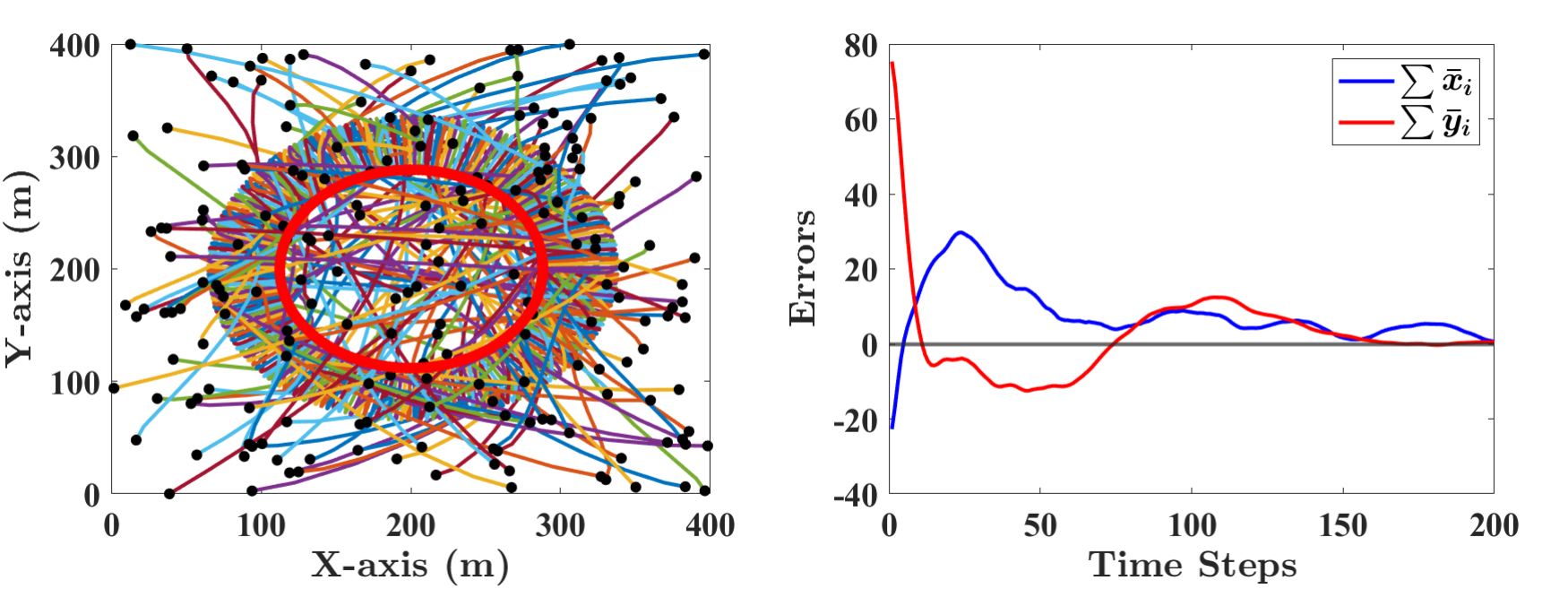}
    \caption{Simulation Results of Robustness.}
    \label{fig:case_R}
    \vspace{-10pt}  
\end{wrapfigure}
\textbf{Robustness:}
In this subsection, we further demonstrate the robustness of the proposed algorithm to communication disruptions. Here, we maintain the same system dynamics and communication topology as in the previous case. However, we randomised the initial points of 200 agents. Additionally, at each time step, we randomly disconnect the communication of half of the robots (600 communication lines). The experimental results are illustrated in Fig. \ref{fig:case_R}.


The graph reveals that, even with half of the communication links disconnected, each robot gradually approaches its Nash equilibrium point. It is important to note that when a robot's communication is disrupted, it still retains the information from the previous time step, allowing it to continue local optimisation based on imperfect historical information. Additionally, when a subset of robots is disconnected, it introduces estimation errors in the global information for the remaining robots, resulting in fluctuations as depicted in the error figure in Fig. \ref{fig:case_R} (right). Even under severe communication challenges, the robots still manage to asymptotically achieve their Nash equilibrium points using the proposed algorithm, thereby illustrating its resilience to communication disruptions.
It is worth noting that the theoretical extreme conditions for convergence are the ergodicity of the continuous-time Markov process and the resulting graph 
containing a spanning tree \cite{you2013consensus}.

\subsection{Real Multi-Robot Systems}
\begin{figure}[htbp]
    \centering
    \includegraphics[width=\hsize]{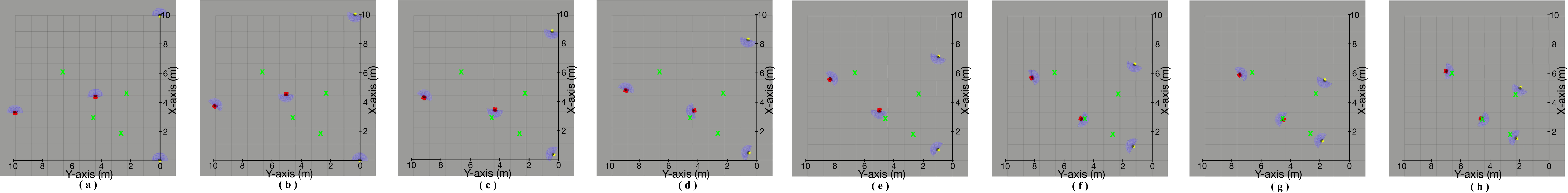}
    \caption{Turtlebots Waffle Pi Simulation Results.}
    \vspace*{-6pt}
    \label{fig:TSim}
\end{figure}

\vspace{-6pt}
\textbf{Software Experiment: }
In this section, we showcase the practical application of our algorithm on real robots, employing the robust physical simulation engine ROS \& GAZEBO. In this scenario, there are two different sizes of robots: two small yellow robots and two large red robots. We use different colours (red and yellow) to represent the distinct system dynamics. The communication graph in this case is shown in Fig. \ref{fig:Ring} (right). The initial points of robots are set as [1,1], [10,1], [5,5], [4,10], and the local targets of each robot are [10, 1], [5,10], [10,2], [3,5]. To make results clearer, robots are designed to turn their head direction along the x-axis once they reach the NE, as shown in, Fig. \ref{fig:TSim}.


The sub-figures (a) to (h) in Fig. \ref{fig:TSim} are the exemplar moments during the seeking process, theoretical optimal Nash equilibrium points are highlighted in green "x" in the figure and a recorded video is available\footnote{\url{https://youtu.be/3GVkfJa07To}}. 
From the figures and video, it becomes evident that the proposed algorithm consistently seeks the Nash equilibrium position for each robot. 
Based on the experimental results, it can be observed that the robots do not continuously move to reach the optimal positions. This is because, at each time step, each robot estimates the positions of all other robots based on the information of its neighbouring robots. As a result of errors in the estimation process, the robots occasionally pause near the target positions. However, our algorithm ensures that the robots ultimately converge to the global optimum. This is because, at the Nash equilibrium, the optimal positions remain stable, and the estimates of the positions of other robots also converge to the optimal points.

\begin{figure}[htbp]
    \centering
    \includegraphics[width=\hsize]{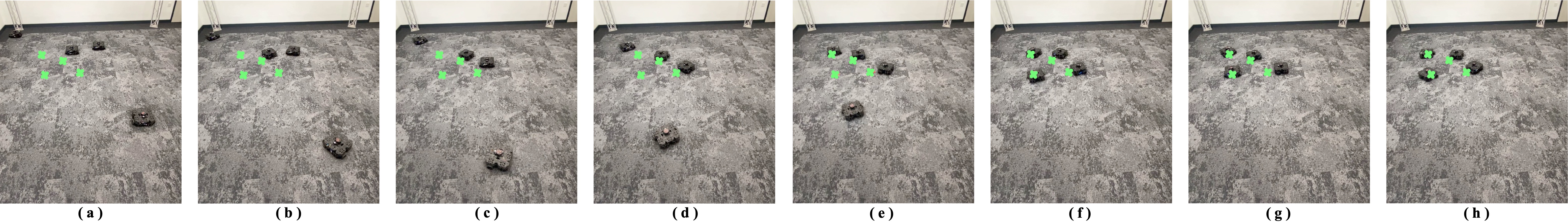}
    \caption{Trajectory of real robots.}
    \vspace{-6pt}
    \label{fig:real}
\end{figure}

\textbf{Real-world Experiment: }
In this subcase, we deploy our algorithm on the real robot platform, Turtlebot waffle pi
, to assess its performance in a practical setting. We introduce four real robots and scatter them randomly across a 3.3 x 3.3-meter laboratory experimental arena. Each robot is assigned a unique local target to simulate diverse goal scenarios. During the experiment, we enforce communication constraints where robots can only communicate with their neighbours at any given time. The robot trajectories are recorded and shown in Fig. \ref{fig:real}. 

\begin{wrapfigure}{r}{0.3\textwidth}
    \centering
    \vspace{-10pt}  
    \includegraphics[width=\hsize]{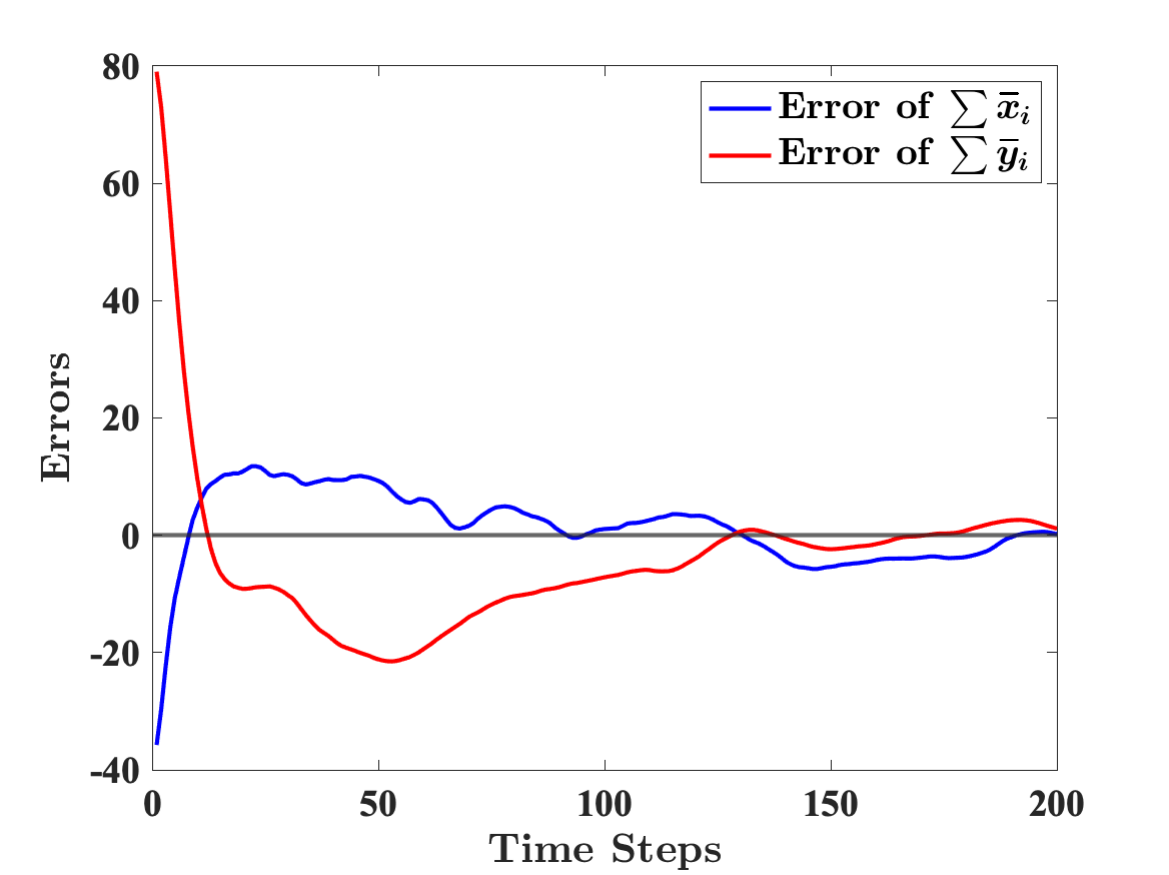}
    \caption{Costs of real robots.}
    \label{fig:Cost}
    \vspace{-10pt}  
\end{wrapfigure}

The sub-figures (a) to (h) in Fig. \ref{fig:real} are the exemplar moments of real robots during the seeking process, and the full video is available\footnote{\url{https://youtu.be/xJStQP3v7Mo}}.  
To illustrate it more clearly, we use red cross tape to represent the Nash equilibrium points, which are highlighted in green in the figures. From the figure, by using the proposed algorithm, we can see that all the robots automatically seek their only optimal Nash equilibrium points. Furthermore, we record the cost of the real robot experiment to show the convergence ability of the proposed algorithm, as shown in Fig. \ref{fig:Cost}.

Fig.~\ref{fig:Cost} shows the value of cost functions among four real robots. While the algorithm successfully guides the multi-robot system toward its Nash equilibrium point, there is a noticeable degradation in tracking performance. This degradation is inherent to the process of seeking a Nash equilibrium, which is not optimised at every individual step. Instead, the optimisation is continuous and iterative, allowing each robot to adjust its strategy based on both its own state and the estimated states of others. Moreover, environmental noise factors—such as slip rates, uneven terrain, or sensor noise—can affect the robots' movements and measurements, introducing additional uncertainty into the system. By monitoring the cost incurred by each robot, our algorithm successfully guides the multi-robot system towards its Nash equilibrium point. This means that, through the process of cooperation, each robot can strategically select actions to maximise its own utility while achieving a balanced state with its adjacent robots. It is noticed that due to hardware limitations in the lab, we only used homogeneous robots to validate the algorithm in this experiment, where the robots' dynamics were uniform; however, our algorithm is scalable to heterogeneous real robots. 
\section{CONCLUSIONS}\label{sec_con}
This paper introduced a novel algorithm that combines output regulation and distributed optimisation to address the challenges associated with achieving Nash equilibrium in fully distributed multi-agent systems. The proposed algorithm aims to guide heterogeneous robots towards their Nash equilibrium points. The proposed approach has been theoretically proven for convergence and has practical applicability, as demonstrated in a real-world multi-robot system.

\newpage
\bibliographystyle{plain}
\bibliography{RSS_ABBR}

\ \ \
\appendix

\section{Proof of Theorem 1}\label{App:A}

\begin{proof}
First, we examine the properties of  $\mathcal{M}_i(k)$: 
\begin{align}    
\begin{aligned}
     \|F_{i}(\xi_i,\hat{v}_i)  -F_{i}(y_i^*,\bar{y}^*)\|^2 \leq 2\|F_{i}(\xi_i,\hat{v}_i)\|^2+2\|F_{i}(y_i^*,\bar{y}^*)\|^2
\end{aligned}
\end{align}
First, we consider the term $\|F_{i}(\xi_i,\hat{v}_i)\|$. There exists a constant $\mathfrak{C}$ such that $\|F_{i}(\xi_i,\hat{v}_i)\|\leq\mathfrak{C}$.
To show this, we have
\begin{align}  
\begin{aligned}
     &\|F_{i}(\xi_i(k),\hat{v}_i(k))\|\\\leq&\|F_{i}(\xi_i(k),\hat{v}_i(k))-F_{i}(\xi_i(k),\bar{\xi}(k))\|+\|F_{i}(\xi_i(k),\bar{\xi}(k))\|\\\leq&L_i\|\hat{v}(k)-\bar{y}(k)\| + \|F_{i}(\xi_i(k),\bar{y}(k))\|
\end{aligned}
\end{align}
Since $\bar{y}(k) \in \Bar{\Omega}$ and $\bar{\xi}(k)$ is a local copy of $\bar{y}(k)$, where $\Bar{\Omega}$ represents the Minkowski mean (or sum) of the sets $\Omega_i$ and $F_i$ is continuous over $\Omega_i\times\Bar{\Omega}$, then it follows that
\begin{align}  
    \|F_{i}(\xi_i(k),\bar{\xi}(k))\| \leq \mathfrak{C}
\end{align}
Furthermore, we also have $\hat{v}_i(k),\bar{\xi}(k) \in \hat{\Omega}$, where $\hat{\Omega} = \cup_i\Omega_i$. Thus,
\begin{align}  
    \|\hat{v}(k)-\bar{\xi}(k)\| \leq \mathfrak{C}
\end{align}
Until now, we can claim that $\|F_{i}(\xi_i,\hat{v}_i)\|$ is bounded.

Also, it is obvious that the last term $\|F_{i}(y_i^*,\bar{y}^*)\|$ is less than the maximum value of $\|F_{i}(y_i^*,\bar{y}^*)\|$ from $i = 1$ to $N$. Therefore, 
\begin{align}  \label{eqn:M}
    \mathcal{M}_i(k) \leq \mathfrak{M} = 2*\mathfrak{C}+2*\mathfrak{Y}
\end{align}


Now, we show the properties of $\mathcal{V}_i(k)$:
\begin{equation}  \label{eqn:N}\small
    \begin{aligned}
        &\left(F_{i}(\xi_i(k),\hat{v}_i(k))  -F_{i}(y_i^*,\bar{y}^*)\right)^T(\xi_i(k) - y_i^*)\\
        =&\left(F_{i}(\xi_i(k),\hat{v}_i(k)) -F_{i}(\xi_i(k),\bar{\xi}(k)) + F_{i}(\xi_i(k),\bar{\xi}(k))-F_{i}(y_i^*,\bar{y}^*)\right)^T(y_i(k) - y_i^*)\\
        =&\left(F_{i}(\xi_i(k),\hat{v}_i(k)) -F_{i}(\xi_i(k),\bar{\xi}(k))\right)^T(\xi_i(k) - y_i^*) +\left(F_{i}(\xi_i(k),\bar{\xi}(k))-F_{i}(y_i^*,\bar{y}^*)\right)^T(\xi_i(k) - y_i^*)\\
        \geq &- 2 \mathfrak{N} L_i\|\hat{v}_i(k)-\bar{\xi}(k)\|+\left(F_{i}(\xi_i(k),\bar{\xi}(k))-F_{i}(y_i^*,\bar{y}^*)\right)^T(\xi_i(k) - y_i^*)
    \end{aligned}
\end{equation}
where $\mathfrak{N} = max_{\xi_i\in\Omega_i}\|\xi_i\|$.

Combining the equations \eqref{eqn:V}, \eqref{eqn:M} and \eqref{eqn:N} yields 
\begin{align}  \label{eqn:V2} 
    \begin{aligned}
        V(k+1)\leq &V(k)+\alpha^2N\mathfrak{M}+4\alpha\mathfrak{N}\sum_{i=1}^N L_i\|\hat{v}_i(k)-\bar{\xi}(k)\|\\&-2\alpha\sum_{i=1}^N\left(F_{i}(\xi_i(k),\bar{\xi}(k))-F_{i}(y_i^*,\bar{y}^*)\right)^T(\xi_i(k) - y_i^*)
    \end{aligned}
\end{align}

Following the Lemma \ref{lemma:robbins},  our remaining task is to demonstrate that $\sum_{i=1}^N L_i\|\hat{v}_i(k)-\bar{\xi}(k)\|<\infty$.
Let $\mathfrak{L} = \max({L_i})$ and $\mathfrak{V} = \theta\beta^k \mathfrak{N}+  N \theta \alpha\mathfrak{C}\sum_{s=1}^k\beta^{k-s}$. Consequently, $\sum_{i=1}^N L_i\|\hat{v}_i(k)-\bar{y}(k)\| \leq N\mathfrak{L}\mathfrak{V}<\infty$.
Therefore, we have 
\begin{equation}  \label{eqn:hatv}\small
     \begin{aligned}
        \hat{v}_i(k)  = &v_i(k+1) - \xi_i(k+1) +\xi_i(k)\\ 
         = &\sum_{j=1}^N a_{ij} v_j(k)\\
         = &\sum_{j=1}^N a_{ij} \left(\sum_{l=1}^N l_{lj} v_l(k-1) + \xi_j(k) - \xi_j(k-1)\right)\\
         = &\sum_{j=1}^N a_{ij} \left(\sum_{l=1}^N l_{lj} \left(\sum_{m=1}^N l_{ml} v_l(k-2) + \xi_l(k-1) - \xi_l(k-2)\right) + \xi_j(k) - \xi_j(k-1)\right)\\
         = &\cdots\\
         =&\sum_{p=1}^N [\Phi(k,0)]_{ip}v_p(0) + \sum_{s=1}^k\left(\sum_{q=1}^N [\Phi(k,s)]_{iq}\left(\xi_q(s) - \xi_q(s-1)\right)\right)
    \end{aligned}
\end{equation}

Meanwhile, it should be noted that 
\begin{align}  \label{eqn:bary}
     \begin{aligned}
        \bar{\xi}(k) = &\bar{\xi}(k-1)+(\bar{\xi}(k)-\bar{\xi}(k-1))\\
        =&\bar{\xi}(0)+\sum_{s=1}^k(\bar{\xi}(s)-\bar{\xi}(s-1))\\
        =&\sum_{p=1}^N\frac{1}{N}v_p(0) + \sum_{s=1}^k\left(\sum_{q=1}^N \frac{1}{N}\left(\xi_q(s) - \xi_q(s-1)\right)\right)
    \end{aligned}
\end{align}

Based on \eqref{eqn:hatv} and \eqref{eqn:bary}, we can get 
    \begin{align*} \label{eqn:v-y}
        &\|\hat{v}_i(k)-\bar{\xi}(k)\| \\ 
        = &\left\|\sum_{p=1}^N [\Phi(k,0)]_{ip}v_p(0)  + \sum_{s=1}^k\left(\sum_{q=1}^N [\Phi(k,s)]_{iq}\left(\xi_q(s) - \xi_q(s-1)\right)\right) \right.\\& \left. - \sum_{p=1}^N\frac{1}{N}v_p(0) + \sum_{s=1}^k\left(\sum_{q=1}^N \frac{1}{N}\left(\xi_q(s) - \xi_q(s-1)\right)\right)\right\|\\\allowdisplaybreaks
        = &\left\|\sum_{p=1}^N \left([\Phi(k,0)]_{ip} - \frac{1}{N}\right)v_p(0)  + \sum_{s=1}^k\left(\sum_{q=1}^N \left([\Phi(k,s)]_{iq} - \frac{1}{N}\right)\left(\xi_q(s) - \xi_q(s-1)\right)\right)\right\|\\\allowdisplaybreaks
        \leq&\left\|\sum_{p=1}^N \left([\Phi(k,0)]_{ip} - \frac{1}{N}\right)v_p(0)\right\| + \left\|\sum_{s=1}^k\left(\sum_{q=1}^N \left([\Phi(k,s)]_{iq} - \frac{1}{N}\right)\left(\xi_q(s) - \xi_q(s-1)\right)\right)\right\|\\\allowdisplaybreaks
        \leq&\sum_{p=1}^N \theta\beta^k\left\|v_p(0)\right\| + \sum_{s=1}^k\sum_{q=1}^N \theta\beta^{k-s}\left\|\left(\xi_q(s) - \xi_q(s-1)\right)\right\|\\
        =&\sum_{p=1}^N \theta\beta^k\left\|v_p(0)\right\| + \sum_{s=1}^k\sum_{q=1}^N \theta\beta^{k-s}\left\|\left(P_{\Omega_q}(\xi_q(s-1) - \alpha F_q(\xi_q(s-1),\hat{v}_q(s-1))) - \xi_q(s-1)\right)\right\|\\
        \leq&\sum_{p=1}^N \theta\beta^k\left\|v_p(0)\right\| + \sum_{s=1}^k\sum_{q=1}^N \theta\beta^{k-s}\alpha\left\|F_q(\xi_q(s-1),\hat{v}_q(s-1))\right\|\\\leq&\theta\beta^k \mathfrak{N}+  N \theta \alpha\mathfrak{C}\sum_{s=1}^k\beta^{k-s}
\end{align*}

It is known that $0<\beta<1$. From the Lemma \ref{lemma:robbins}
and let $\gamma_k = 1$, the following inequality function can be obtained
\begin{align}  
\sum_{k=1}^N\sum_{s=1}^k\beta^{k-s} \leq \infty
\end{align}
which implies
\begin{align}  
    \begin{aligned}
        \sum_{i=1}^N L_i\|\hat{v}_i(k)-\bar{\xi}(k)\|<\infty
    \end{aligned}
\end{align}
This completes the proof.
\end{proof}

\section{Proof of Theorem 2}\label{App:B}

\begin{proof}
    We first demonstrate that, for any bounded $\Delta_i(k) $, we have 
    \begin{align}  \label{eqn: input to output stability}
		\limsup_{k\rightarrow \infty} \| e_i(k) \| \leq  \alpha  \limsup_{k\rightarrow \infty} \|\Delta_i(k)\|.
	\end{align}
	Encapsulating \eqref{eqn: linear matrix align} into a matrix form leads to 
	\begin{align}  \label{eqn: matrix align}
		\left[\begin{array}{cc}
			A_i-I & B_i\\
				     C_i & 0
		\end{array} \right]   \left[\begin{array}{c}
			\Psi_i  \\
			G_i
		\end{array} \right]=  \left[\begin{array}{c}
			0 \\
			I
		\end{array} \right].
	\end{align}
	Denoting $\mathcal O_i = 	\left[\begin{array}{cc}
		A_i-I & B_i\\
			     C_i & 0 
	\end{array} \right]  $ and $T_i = \left[\begin{array}{c}
		\Psi_i \\
		G_i
	\end{array} \right]$, by leveraging the property of Kronecker product, $\operatorname{vec}\left(\mathcal O_i T_i I \right)=\left(I \otimes \mathcal O_i\right) \operatorname{vec}(T_i)$, we can obtain a standard linear algebraic equation 
	\begin{align}  \label{key}
		\left(I \otimes \mathcal O_i\right) \operatorname{vec}(T_i) =\operatorname{vec} \left(\left[\begin{array}{c}
			0  \\
			I
		\end{array} \right]  \right) 
	\end{align}
of which the solvability is guaranteed under \eqref{eqn: rank condition} in Assumption~\ref{asm: rank}. 
	
	For notation convenience, we denote $A_{i,c} = A_i-B_iK_i$ and $B_{i,c} = -\Psi_i$. Then, we have 
\begin{align}   \label{eqn: closed-loop error dynamics 2}\begin{aligned}
		\bar x_i(k+1)  & =  A_{i,c}\bar x_i (k) +B_{i,c} \Delta_i (k) .
	\end{aligned}
\end{align}
Recursively iterating \eqref{eqn: closed-loop error dynamics 2} results in 
\begin{align}  \label{eqn: denition of time index}
	\bar x_i(k) = A_{i,c}^k\bar x_i(0) +\sum_{j=0}^{k-1} A_{i,c}^{k-j-1}B_{i,c} \Delta_i(j).
\end{align}
Hence, we have
\begin{align}  \label{eqn: error equation 1}
	e_i(k) = C_i\bar x(k) =C_iA_{i,c}^k\bar x_i(0) - \sum_{j=0}^{k-1} A_{i,c}^{k-j-1} \Delta_i(j)
\end{align}
where $C_i\Psi_i - I =0 $ has been used. Because $A_{i,c} $ is Schur, we have $\lim_{k\rightarrow \infty} C_iA_{i,c}^k\bar x_i(0) = 0 $. In Section \ref{parta}, the convergence of reference generator has been established, which implies $\Delta_i$ converges to zero as $k\rightarrow \infty$. Denote $\varpi_i:= \limsup_{k\rightarrow \infty} \| \Delta_i(k)\| $. Then, for any small constant $\epsilon>0$, there exists a positive time index $\zeta>0$ such that 
\begin{align}  \label{eqn: bounds of gradient term}
	\| \Delta_i(k)\| < \varpi_i+\epsilon, \  \forall k>\zeta. 
\end{align}
Based on the time index $\zeta$, the second term in \eqref{eqn: error equation 1} can be separated into two parts, written as 
\begin{align}  \label{eqn: second term separation}
	\sum_{j=0}^{k-1} A_{i,c}^{k-j-1} \Delta_i(j) = \sum_{j=0}^{\zeta} A_{i,c}^{k-j-1} \Delta_i(j)+\sum_{j=\zeta+1}^{k-1} A_{i,c}^{k-j-1} \Delta_i(j).
\end{align}
Taking the Euclidean norm of \eqref{eqn: second term separation} and invoking \eqref{eqn: bounds of gradient term}:
\begin{align}  \label{eqn: bounded second term}\begin{aligned}
	&\bigg\|\sum_{j=0}^{k-1} A_{i,c}^{k-j-1} \Delta_i(j)\bigg\| =  \bigg\| \sum_{j=0}^{\zeta} A_{i,c}^{k-j-1} \Delta_i (j) +\sum_{j=\zeta+1}^{k-1} A_{i,c}^{k-j-1} \Delta_i(j) \bigg\| \\
	& \leq \big\| A_{i,c}^{k-\zeta-1} \big \| \bigg\| \sum_{j=0}^{\zeta} A_{i,c}^{\zeta-j} \Delta_i(j) \bigg \| + (\varpi_i +\epsilon) \bigg\| \sum_{j=\zeta+1}^{k-1} A_{i,c}^{k-j-1}  \bigg\| . 
	\end{aligned}
\end{align}
Therefore, we have 
\begin{align}  \label{eqn: bound of e}
	\limsup_{k \rightarrow \infty} \|e_i(k) \| \leq \frac{1}{1-\|A_{i,c}\|}\left(\varpi_i +\varepsilon\right)
\end{align}
where the following two results have been applied
\begin{align}  
	\sum_{j=\zeta+1}^{t-1}\|A_{i,c}\|^{t-1-j}=\frac{1-\|A_{i,c}\|^{t-\zeta}}{1-\|A_{i,c}\|}<\frac{1}{1-\|A_{i,c}\|}
\end{align}
\begin{align}  
	\lim_{k\rightarrow \infty} \big\| A_{i,c}^{k-\zeta-1} \big \|  = 0 .
\end{align}
As $\epsilon $ can be set arbitrarily small, it follows from \eqref{eqn: bound of e} that 
\begin{align}  \label{eqn: input to output stability 2}
		\limsup_{k\rightarrow \infty} \| e_i(k) \| \leq \alpha \limsup_{k\rightarrow \infty} \|\Delta_i(k)\|.
\end{align}
where $\alpha =\frac{1}{1-\|A_{i,c}\|} $. 

Finally, let $\tilde x_i(k) = x_i(k)-\Psi_i y_i^*$. Then, we have 
\begin{align}  \begin{aligned} 
	\tilde x_i(k+1) = & A_ix_i(k) + B_i [-K_ix_i(k) +(G_i+ K_i\Psi_i) \xi_i (k) ] - \Psi_i y_i^* \\
	 = & (A_i-B_iK_i) \tilde x_i(k) + B_i(G_i+K_i\Psi_i) (\xi_i(k) - y_i^*).
	\end{aligned}
\end{align}
It follows from Theorem~\ref{thm: 1} and \eqref{eqn: input to output stability 2} that $\xi_i(k) $ converges to $y_i^*$. Thus, we can conclude the convergence of the proposed algorithm \eqref{solution:B} by treating $B_i(G_i+K_i\Psi_i) (\xi_i(k) - y^*)$ as $\Delta_i(k)$ in \eqref{eqn: input to output stability 2}. 
\end{proof}

\end{document}